%% file: main.tex
\documentclass[11pt,a4paper]{article}
\usepackage[hyperref]{acl2019}
\usepackage{times}
\usepackage{latexsym}

\usepackage{url}

\aclfinalcopy 
\def\aclpaperid{813} 

\usepackage{soul}
\usepackage[utf8]{inputenc}
\usepackage{graphicx}
\usepackage{booktabs}
\usepackage{xcolor}
\usepackage[vlined,ruled,linesnumbered]{algorithm2e}
\usepackage{algorithmic}
\usepackage{longtable}
\usepackage{mathtools}
\usepackage{CJK}
\usepackage{multirow}
\usepackage{multicol}
\usepackage{amsmath,amssymb,amsfonts}
\usepackage{bm}
\usepackage[all]{nowidow}

\title{Exploiting Entity BIO Tag Embeddings and Multi-task Learning for Relation Extraction with Imbalanced Data}

\author{
	Wei Ye$^{1}$*\thanks{* indicates equal contribution.}$^\text{\textdagger}$\thanks{$^\text{\textdagger}$ Corresponding author.}, Bo Li$^{1,3}$*, Rui Xie$^{1,2}$, \textbf{Zhonghao Sheng}$^{1,2}$,\\
	\textbf{Long Chen}$^{1,2}$ \and \textbf{Shikun Zhang}$^{1}$ \\
	$^1$National Engineering Research Center for Software Engineering, Peking University\\
	$^2$School of Software and Microelectronics, Peking University\\
	$^3$Automation Dept, Beijing University of Posts and Telecommunications.\\
	wye@pku.edu.cn, deepblue.lb@gmail.com, \\
	\{ruixie, zhonghao.sheng, clcmlxl, zhangsk\}@pku.edu.cn
}

\renewcommand\footnotemark{}
\date{}

\begin{document}
\begin{CJK}{UTF8}{gbsn}
\maketitle
\begin{abstract}
    In practical scenario, relation extraction needs to first identify entity pairs that have relation and then assign a correct relation class. However, the number of non-relation entity pairs in context (negative instances) usually far exceeds the others (positive instances), which negatively affects a model's performance. To mitigate this problem, we propose a multi-task architecture which jointly trains a model to perform relation identification with cross-entropy loss and relation classification with ranking loss. Meanwhile, we observe that a sentence may have multiple entities and relation mentions, and the patterns in which the entities appear in a sentence may contain useful semantic information that can be utilized to distinguish between positive and negative instances. Thus we further incorporate the embeddings of character-wise/word-wise BIO tag from the named entity recognition task into character/word embeddings to enrich the input representation. Experiment results show that our proposed approach can significantly improve the performance of a baseline model with more than 10\% absolute increase in F1-score, and outperform the state-of-the-art models on ACE 2005 Chinese and English corpus. Moreover, BIO tag embeddings are particularly effective and can be used to improve other models as well. 
\end{abstract}
\input{Introduction}
\input{Approach}
\input{Experiment}

\input{Literature}

\input{Conclusion}

\section*{Acknowledgments}
We would like to thank Handan Institute of Innovation, Peking University for their support of our work.

\bibliographystyle{acl_natbib}
\bibliography{refer}

\end{CJK}
\end{document}

%% file: Introduction.tex
\section{Introduction}\label{sec:introduction}

Relation extraction, which aims to extract semantic relations from a given instance---entity pair and the corresponding text in context, is an important and challenging task in information extraction. It serves as a step stone for many downstream tasks such as question answering and knowledge graph construction.

Traditionally, researchers mainly use either feature-based methods \cite{kambhatla2004combining,boschee2005automatic,guodong2005exploring,Jiang2007systematic,chan2010background,sun2011semi,Nguyen2014Regularization} or kernel-based methods \cite{Zelenko2003Kernel,Culotta2004Dependency,bunescu2005shortest,mooney2006subsequence,zhang2006exploring,zhou2007tree,giuliano2007fbk,qian2008exploiting,Nguyen2009Structures,Sun2014tree} for relation extraction, which tend to heavily rely on handcraft features and existing natural language processing (NLP) tools. Recently, deep learning models, including convolutional neural network (CNN) \cite{liu2013convolution,zeng2014relation,nguyen2015relation,zeng2015distant,santos2015classifying,lin2016neural} and recurrent neural network (RNN) \cite{miwa2016end,zhou2016attention,she2018distant} w/o variants of attention mechanism have been widely applied to relation extraction and achieved remarkable success.

The relation extraction task can be divided into two steps: determining which pair of entities in a given sentence has relation, and assigning a correct relation class to the identified entity pair. We define these two steps as two related tasks: \textbf{Relation Identification} and \textbf{Relation Classification}. If one only needs to categorize the given entities that are guaranteed to have some expected relation, then relation extraction is reduced to relation classification \cite{nguyen2015relation}. 

One variation of relation classification is the introduction of a new artificial relation class ``Other.'' If the number of non-relation entity pairs in context (\textbf{negative instances}) in the dataset is comparable to the number of entity pairs that have relation in context (\textbf{positive instances}), then the non-relation pairs can be treated as having the relation class Other. 

Strictly speaking, most existing studies of relation extraction treat the task as relation classification. However, relation extraction often comes with an extremely imbalanced dataset where the number of non-relation entity pairs far exceeds the others, making it a more challenging yet more practical task than relation classification. For example, after filtering out those entity pairs whose entity type combination has never appeared in the Chinese corpus of ACE 2005, there are still more than 200,000 entity pairs left, in which the positive/negative instance ratio is about 1:20. In this paper, we focus on the relation extraction task with an imbalanced corpus, and adopt multi-task learning paradigm to mitigate the data imbalance problem.

Only a few studies have considered the negative effect of having too many negative instances. \citet{nguyen2015relation} proposed using CNN with filters of multiple window sizes. \citet{santos2015classifying} focused on learning the common features of the positive instances by computing only the scores of the relation classes excluding the class Other, and proposed using a pairwise ranking loss. We have also adopted these methods in our approach.

For relation classification, the prediction error can be categorized into three types: 1) false negative---predicting a positive instance to be negative; 2) false positive---predicting a negative instance to be positive; 3) wrong relation class--- predicting a positive instance to be positive yet assigning a wrong relation class. After training a baseline model to perform relation classification on the extremely imbalanced ACE 2005 Chinese corpus and dissecting its prediction errors, we find that the proportion of these three types of error are $30.20\%$, $62.80\%$ and $7.00\%$ respectively. It is conceivable that to improve a model's performance on such corpus, it is best to focus on telling positive and negative instances apart.


Since the negative instances may not have much in common, distinguishing between positive and negative instances is much more challenging than only classifying positive instances into a correct class. Moreover, the total number of positive instances combined is more comparable to the number of negative instances than positive instances of any individual relation class alone. Based on these rationales, we propose to jointly train a model to do another binary classification task---relation identification---alongside relation classification to mitigate the data imbalance problem.

Another facet that most existing studies fail to consider is that there may be multiple relation mentions in a given sentence if it contains multiple entities. In the Chinese corpus of ACE 2005, there are 4.9 entities and 1.34 relation mentions in a sentence on average. The patterns in which these entities appear in the sentence can provide useful semantic information to distinguish between positive and negative instances. Therefore, we exploit the character-wise/word-wise BIO (Beginning, Inside, Outside) tag used in the named entity recognition (NER) task to enrich the input representation. The details of our approach will be presented in Section ~\ref{sec:Proposed Method}.

We conducted extensive experiments on ACE 2005 Chinese and English corpus. Results show that both the novel multi-task architecture and the incorporation of BIO tag embeddings can improve the performance, and the model equipped with both achieves the highest F1-score, significantly outperforming the state-of-the-art models. Analysis of the results indicates that our proposed approach can successfully address the problem of having a large number of negative instances.

To summarize, we make the following contributions in this paper:

\begin{enumerate}
	\item We propose a multi-task architecture which jointly trains a model to perform relation identification with cross-entropy loss and relation classification task with ranking loss, which can successfully mitigate the negative effect of having too many negative instances.
	
	\item We incorporate the embeddings of character-wise/word-wise BIO tag from NER task to enrich the input representation, which proves to be very effective not only for our model but for other models as well. We argue that BIO tag embeddings could be a general part of character/word representation, just like the entity position embeddings \cite{zeng2014relation} that many researchers would use in recent years.
\end{enumerate}

%% file: Approach.tex
\section{Proposed Approach}\label{sec:Proposed Method}

\begin{figure}[ht]
	\centering
	\includegraphics[width=\columnwidth]{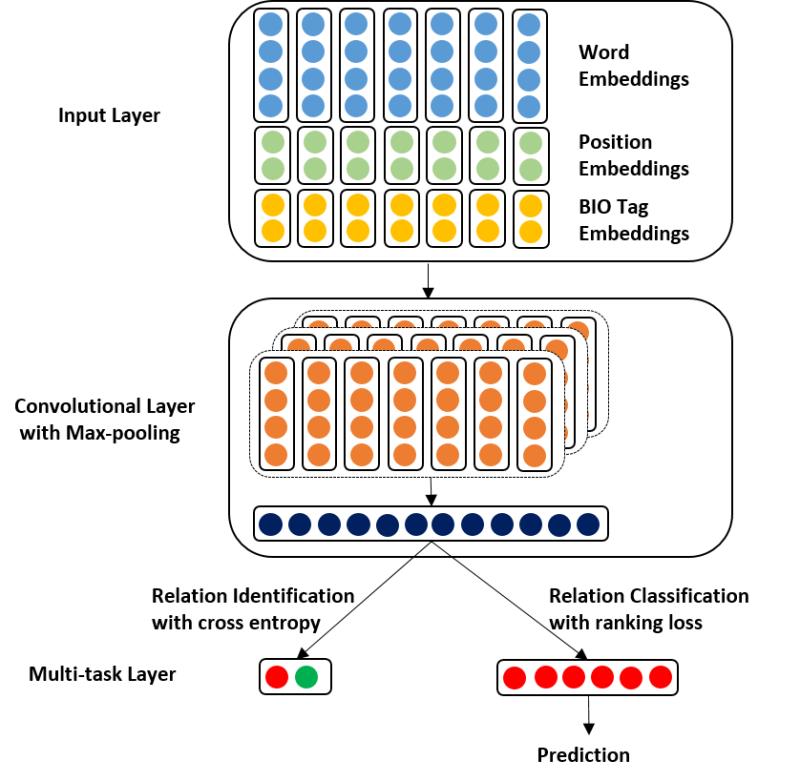}
	\caption{The overall multi-task architecture. To demonstrate, there are three window sizes for filters in the convolutional layer, as denoted by the three-layer stack; for each window size there are four filters, as denoted by the number of rows in each layer. Max-pooling is applied to each row in each layer of the stack, and the dimension of the output is equal to the total number of filters.}
	\label{overview_fig}
\end{figure}

We have designed a novel multi-task architecture which combines two related tasks: 1) relation identification, which is a binary classification problem to determine whether a given entity pair has relation; 2) relation classification, which is a multiple classification problem to determine the relation class. Figure \ref{overview_fig} shows the overall architecture.

\begin{figure*}[t]
	\centering
	\includegraphics[width=\linewidth]{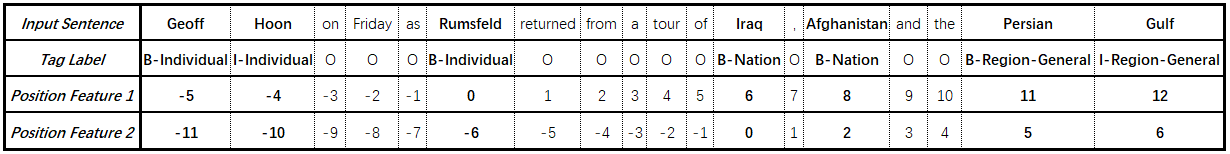}
	\caption{Illustration of BIO tag information and positional information for a given instance. In this example, there are five entities in the input sentence, and the target entities are the second and the third.}
	\label{tag}
\end{figure*}

Three are three main parts in the architecture:
\begin{itemize}
	\item \textbf{Input Layer} Given an input sentence $x$ of $n$ words\footnote{We use character-wise model for Chinese corpus and word-wise model for English corpus. For simplicity sake, we use ``word'' to denote either an English word or a Chinese character to present our model.} $\{x_1, x_2,..., x_n\}$ with $m$ entities $\{e_1, e_2, ..., e_m\}$ where $e_i \in \bm{x}$, and two target entities $e_{t1}, e_{t2} \in \{e_1, e_2, ..., e_m\}$, the input layer transforms the sentence into a matrix $\bm{X}$, which includes word embeddings, position embeddings and BIO tag embeddings of each word.
	
	\item \textbf{Convolutional Layer with Max-pooling} Following the input layer is a convolutional layer that extracts high-level features, with filters (convolution kernels) of multiple window sizes \cite{nguyen2015relation}. Then max-pooling is applied to each feature map to reduce dimensionality. 
	
	\item \textbf{Multi-task Layer} In the multi-task layer, the model jointly learns the relation identification task using cross-entropy loss and the relation classification task using ranking loss.
	
\end{itemize}

\subsection{Input Layer}

\begin{itemize}
	\item \textbf{Word Embeddings}
	We use word embeddings with random initialization for each word in the input sentence. The dimension of word embeddings is $d^w$.
	
	\item \textbf{Position Embeddings}
	We also employ position embeddings to encode the relative distance between each word and the two target entities in the sentence. We believe that more useful information regarding the relation is hidden in the words closer to the target entities. The dimension of position embeddings is $d^p$.
	
	\item \textbf{BIO Tag Embeddings}
	Since an input sentence often contains more than two entities, we utilize the BIO tag information of entities to enrich the representation of the input. More specifically, for each word in the input sentence, if the word is part of an entity, we use the entity type $T$ to label the start of the entity as $B_T$, and label the rest of the entity as $B_I$. If the word is not part of an entity, then we label the word as $O$. The dimension of BIO tag embeddings is $d^t$.
\end{itemize}

After concatenating all three embeddings together for each word, we transform a sentence into a matrix $\bm{X}=[\bm{w_1}, \bm{w_2}, ..., \bm{w_n}]$ as the input representation, where the column vector $\bm{w_i}\in \mathbb{R}^{d^w+2*d^p+d^t}$. Figure \ref{tag} illustrates how to derive position embeddings and BIO tag embeddings.

\subsection{Convolutional Layer with Multi-Sized Window Kernels}
Next, the matrix $\bm{X}$ is fed into the convolutional layer to extract high-level features. A filter with window size $k$ can be denoted as $\bm{F}=[\bm{f_1},\bm{f_2},.., \bm{f_k}]$, where the column vector $\bm{f_i} \in \mathbb{R}^{d^w+2*d^p+d^t}$. Apply the convolution operation on the two matrices $\bm{X}$ and $\bm{F}$, and we get a score sequence $T=\{t_1, t_2,..., t_{n-k+1}\}$:

\begin{equation}
t_i=g(\sum_{j=0}^{k-1}\bm{f_{j+1}}^T\bm{w_{j+i}}+b)
\label{eq:convolution}
\end{equation}
where $g$ is a non-linear function and $b$ is bias. 

In our experiments, we apply zero-paddings during the convolution operation, so that the score sequence has the same length as the input sequence, which is $n$, instead of $n-k+1$ if we apply Equation \ref{eq:convolution} which assumes no padding.

There are multiple filters with different window sizes in the convolutional layer. Then max-pooling is applied to the outputted feature map of each filter. Eventually the input sentence $x$ is represented as a column vector $\bm{r}$ with a dimension that is equal to the total number of filters.

\subsection{Multi-Task Layer}

\begin{itemize}
	\item \textbf{Relation Identification with Cross-entropy Loss}
	For the binary classification task of relation identification we use cross-entropy loss. Positive instances are labelled ``1'' and negative instances ``0.''
	
	If $p$ is the one-hot true distribution over all classes $C=\{c\}$ and $q$ is the distribution a model predicts, then the cross-entropy loss of a given instance can be defined as follows:
	
	\begin{equation}
	H(p,q)=-\sum_{c\in C}p(c)log(q(c))
	\end{equation}
	
	So the loss of this task can be defined as: 
	\begin{equation}
	loss_1=-\sum(p(1)log(q(1)) + p(0)log(q(0)))
	\end{equation}
	
	\item \textbf{Relation Classification with Ranking Loss}
	For the multiple classification task of relation classification, we use the pairwise ranking loss proposed by \cite{santos2015classifying}.
	
	Given the sentence representation $\bm{r}$, the score for class $c$ is computed as:
	\begin{equation}
	s_c=\bm{r}^T[\bm{W}^{classes}]_c
	\end{equation}
	where $\bm{W}^{classes}$ is a matrix to be learned, whose number of columns is equal to the number of classes. $\bm{W}^{classes}_c$ is a column vector corresponding to class $c$, whose dimension is equal to that of $\bm{r}$.
	
	For each instance, the input sentence $x$ has a correct class label $y^+$ and incorrect ones $y^-$. Let $s_{y^+}$ and $s_{y^-}$ be the scores for $y^+$ and $y^-$ respectively, then the ranking loss can be computed by the following two equations:
	\begin{equation}
	L^+=log(1+exp(\gamma(m^+ - s_{y^+})))
	\end{equation}
	\vspace{-11pt}
	\begin{equation}	
	L^-=log(1+exp(\gamma(m^- + s_{y^-})))
	\end{equation}
	where $m^+$ and $m^-$ are margins and $\gamma$ is a scaling factor.	$L^+$ decreases as the score $s_{y^+}$ increases, and is close to zero when $s_{y^+} > m^+$, which encourages the network to give a score greater than $m^+$ for the correct class. Similarly, $L^-$ decreases as the score $s_{y^-}$ decreases, and is close to zero when $s_{y^-} < -m^-$, which encourages the network to give scores smaller than $-m^-$ for incorrect classes. 
	
	For the class Other, only $L^-$ is calculated to penalize the incorrect prediction. And following \cite{santos2015classifying}, we only choose the class with the highest score among all incorrect classes as the one to perform a training step. Then we optimize the pairwise ranking loss function:
	\begin{equation}
	loss_2=\sum(L^+ + L^-)
	\end{equation}
	
\end{itemize}

The total loss function for multi-task training is:
\begin{equation}
L = \alpha \cdot loss_1 + \beta \cdot loss_2
\end{equation}
where $\alpha$ and $\beta$ are weights of the two losses. In our experiments, we find that $\alpha = \beta$ yields the best result. 

\subsection{Prediction}

We only use the class score $s_c$ in the multiple classification task to make predictions, while the binary classification task is only used for optimizing the network parameters.

Given an instance, the prediction $P$ is made by:
\begin{equation}
P=\left\{
\begin{array}{rcl}
\mathop{\arg\max} \limits_{c} (s_c)     &      & {max(s_c) \geq \theta}\\
Other     &      & {max(s_c) < \theta}\\
\end{array} \right. 
\end{equation}
where $\theta$ is a threshold. The relation in an instance is predicted as the class Other if the score $s_c$ is less than $\theta$ for every class $c$. Otherwise, we choose the class with the highest score as the prediction.

%% file: Experiment.tex
\section{Experiments and Results}\label{sec:Experiment and Results}
\subsection{Data Preparation}
We use both the Chinese and English corpus of ACE 2005 to evaluate our proposed approach. Only positive instances have been annotated in the dataset. To extract negative instances, we need to enumerate every entity pair in a sentence. 

We consider two approaches: one considers the direction of relation while the other does not. For the first approach, we extract only one instance for any pair of entities $e_1$, $e_2$ in a sentence $x$ regardless of direction. Those instances that have been annotated, regardless of direction, are positive instances, and the rest are negative instances. A trained model only needs to determine whether an entity pair has relation. For the second approach, we extract two instances for any pair of entities in a sentence, with the two entities in different orders. Since at most one of the two instances has been annotated to be positive instances, we treat the other one and those neither of which are annotated to be negative instances. A trained model will additionally need to identify head entity and tail entity in a relation, which is considerably harder. 

After extracting negative instances, we further filtered out those instances whose entity type combination has never appeared in a relation mention. Then we added the remaining negative instances to the positive instances to complete data preparation.
 
We adopted the first approach to extract negative instances from the English corpus of ACE 2005, and ended up with 71,895 total instances after filtering, among which 64,989 are negative instances. The positive/negative instance ratio is about 1:9.4. 

We adopted the second approach to extract negative instances from the Chinese corpus of ACE 2005, and ended up with 215,117 total instances after filtering, among which 205,800 of them are negative instances. The positive/negative instance ratio is about 1:20. 


\subsection{Experiment Settings}
\subsubsection{Embeddings}
In our approach, we use three kinds of embeddings, namely word embeddings, position embeddings and BIO tag embeddings. They are all randomly initialized, and are adjusted during training. The dimensions of these three embeddings are 200, 50 and 50 respectively.

\subsubsection{Hyper-parameters}
The number of filters in the convolutional layer is 64, and the window size of filters ranges from 4 to 10. The fully connected layer to calculate class scores has 128 hidden units with a dropout rate of 0.2. The batch size is 256. The neural networks are trained using the RMSprop optimizer with the learning rate $\alpha$ set to 0.001. 

As for the parameters in the pairwise ranking loss, for the English corpus, we set $m^+$ to 2.5, $m^-$ to 0.5, $\gamma$ to 2 and $\theta$ to 0; for the Chinese corpus, we set $m^+$ to 4.5, $m^-$ to -0.5, $\gamma$ to 2 and $\theta$ to 1. The cross-entropy loss and the pairwise ranking loss in multi-task learning are equally weighted.

\subsection{Experiment Results}

We use five-fold cross-validation to reduce the randomness in the experiment results. The precision ($\textbf{P}$), recall ($\textbf{R}$) and F1-score ($\textbf{F1}$) of the positive instances are used as evaluation metrics. 


We compare several variants of our proposed models with the state-of-the-art models on the English and Chinese corpus of ACE 2005 respectively. Variants of our models are:

\begin{itemize}
	\item Baseline: a model that uses CNN with filters of multiple window sizes and only performs the relation classification task using the pairwise ranking loss. The baseline model is motivated by \citet{santos2015classifying} and \citet{nguyen2015relation}.
	\item Baseline+Tag: baseline model with BIO tag embeddings.
	\item Baseline+MTL: baseline model that performs relation identification using cross-entropy loss in addition to relation classification.
	\item Baseline+MTL+Tag, baseline model that adopts both multi-tasking learning and BIO tag embeddings.
\end{itemize}

For the English corpus, we choose SPTree \cite{miwa2016end} and Walk-based Model \cite{christopoulou2019walk} for comparison. Since the data preparation is similar, we directly report the results from the original papers. The experiment results are summarized in Table \ref{experiment1_en}.

For the Chinese corpus, we choose PCNN \cite{zeng2015distant} and Eatt-BiGRU \cite{qin2017designing} for comparison. We re-implemented these two models, and the experiment results are summarized in Table \ref{experiment1_cn}.

\begin{table}[h]
	\renewcommand
	\renewcommand{\multirowsetup}{\centering}
	\centering
	\setlength{\tabcolsep}{3mm}\begin{tabular}{|c|c|c|c|}
		\hline
		\textbf{Model}&\textbf{P\%}&\textbf{R\%}&\textbf{F1\%}
		\\\hline
		SPTree&70.1&61.2&65.3
		\\\hline
		Walk-based&69.7&59.5&64.2
		\\\hline
		Baseline&58.8&57.3&57.2
		\\\hline
		\textbf{Baseline+Tag}&61.3&76.7&\textbf{67.4}
		\\\hline
		Baseline+MTL&63.8&56.1&59.5
		\\\hline
		\textbf{Baseline+MTL+Tag}&66.5&71.8&\textbf{68.9}
		\\\hline
	\end{tabular}
	\caption{Comparison between our model and the state-of-the-art models using ACE 2005 English corpus. F1-scores higher than the state-of-the-art are in bold.}
	\label{experiment1_en}
\end{table}

\begin{table}[h]
	\renewcommand
	\renewcommand{\multirowsetup}{\centering}
	\centering
	\setlength{\tabcolsep}{3mm}\begin{tabular}{|c|c|c|c|}
		\hline
		\textbf{Model}&\textbf{P\%}&\textbf{R\%}&\textbf{F1\%}
		\\\hline
		PCNN&54.4&42.1&46.1
		\\\hline
		Eatt-BiGRU&57.8&49.7&52.0
		\\\hline
		Baseline&48.5&57.1&51.7
		\\\hline
		\textbf{Baseline+Tag}&61.8&62.7&\textbf{61.4}
		\\\hline
		\textbf{Baseline+MTL}&56.7&52.9&\textbf{53.8}
		\\\hline
		\textbf{Baseline+MTL+Tag}&61.3&65.8&\textbf{62.9}
		\\\hline
	\end{tabular}
	\caption{Comparison between our model and the state-of-the-art models using ACE 2005 Chinese corpus. F1-scores higher than the state-of-the-art are in bold.}
	\label{experiment1_cn}
\end{table}

From Table \ref{experiment1_en} and Table \ref{experiment1_cn}, we can see:
\begin{enumerate}
	\item Both BIO tag embeddings and multi-task learning can improve the performance of the baseline model. 
	
	\item Baseline+Tag can outperform the state-of-the-art models on both the Chinese and English corpus. Compared to the baseline model, BIO tag embeddings lead to an absolute increase of about 10\% in F1-score, which indicates that BIO tag embeddings are very effective.	
	
	\item Multi-task learning can yield further improvement in addition to BIO tag embeddings: Baseline+MTL+Tag achieves the highest F1-score on both corpora.
\end{enumerate}

\subsection{Analysis}
\subsubsection{Effectiveness of BIO Tag Embeddings}

To further investigate the effectiveness of BIO tag embeddings, we incorporated these embeddings into PCNN \cite{zeng2015distant} and Eatt-BiGRU \cite{qin2017designing} to form two new models: PCNN+Tag and East-BiGRU+Tag, and evaluated their performance using the Chinese corpus of ACE 2005. The results are summarized in Table \ref{experiment_tag}.

\begin{table}[h]
	\renewcommand
	\renewcommand{\multirowsetup}{\centering}
	\centering
	\setlength{\tabcolsep}{3mm}\begin{tabular}{|c|c|c|c|}
		\hline
		\textbf{Model}&\textbf{P\%}&\textbf{R\%}&\textbf{F1\%}
		\\\hline	
		PCNN+Tag&74.3&50.4&58.2
		\\\hline
		Eatt-BiGRU+Tag&67.8&56.4&61.1
		\\\hline
	\end{tabular}
	\caption{Evaluation of state-of-the-art models with BIO Tag embeddings using ACE 2005 Chinese corpus.}
	\label{experiment_tag}
\end{table}

Compare Table \ref{experiment_tag} with Table \ref{experiment1_cn}, and we can see that thanks to BIO tag embeddings, the F1-score of PCNN increases from 46.1\% to 58.2\%, while the F1-score of Eatt-BiGRU increases from 52.0\% to 61.1\%. Such significant improvement is consistent with that on the baseline model and further attests to the effectiveness of BIO tag embeddings. We believe that BIO tag embeddings could be used as a general part of character/word representation for other models and potentially other tasks as well.

\subsubsection{Effect of Positive/Negative Instance Ratio}

\begin{figure}[h]
	\centering
	\includegraphics[width=\linewidth]{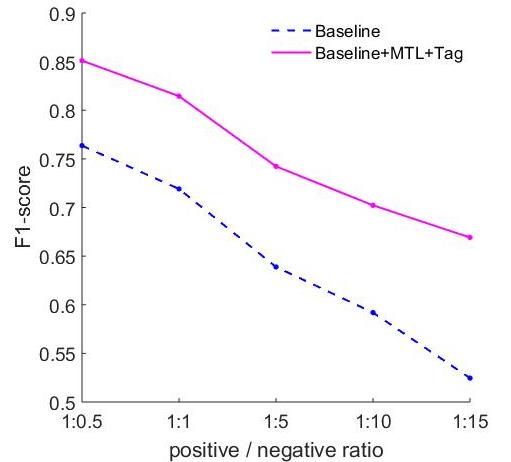}
	\caption{Effect of positive/negative instance ratio on F1-score.}
	\label{experiment2}
\end{figure}

To see how our approach would perform as the degree of data imbalance varies, we used the same random seed to sample negative instances extracted from the Chinese corpus of ACE 2005 to add to the positive instances with different negative/positive instance ratios of 1:0.5, 1:1, 1:5, 1:10 and 1:15. Then we trained and evaluated two models: Baseline and Baseline+MTL+Tag. The results are shown in Figure \ref{experiment2}.

\begin{table*}[t]
	\renewcommand
	\renewcommand{\multirowsetup}{\centering}
	\centering
	\setlength{\tabcolsep}{3mm}\begin{tabular}{|c|c|c|c|c|c|}
		\hline
		\textbf{Model}&\textbf{RI}&\textbf{Loss Function in RC}&\textbf{P\%}&\textbf{R\%}&\textbf{F1\%}
		\\\hline
		Baseline+Tag&$\texttimes$&Ranking Loss&61.8&62.7&61.4
		\\\hline
		Baseline+Tag&$\texttimes$&Cross-entropy Loss&67.7&57.8&61.5		
		\\\hline
		Baseline+Tag&$\texttimes$&Cross-entropy Loss + Ranking Loss&63.2&62.1&61.7
		\\\hline
		Baseline+MTL+Tag&$\checkmark$&Ranking Loss&61.3&65.8&62.9
		\\\hline
		Baseline+MTL+Tag&$\checkmark$&Cross-entropy Loss&61.6&62.0&62.0
		\\\hline
	\end{tabular}
	\caption{Evaluating the effect of the loss function used in relation classification w/o multi-tasking using ACE 2005 Chinese corpus. RC stands for relation classification and RI stands for relation identification.}
	\label{experiment_loss}
\end{table*}

As shown in Figure \ref{experiment2}, the performance drops for both models in terms of F1-score as the positive/negative instance ratio decreases. Yet, as the data become more imbalanced, the gap between the performances of Baseline+MTL+Tag and Baseline widens. This indicates that our proposed approach is more useful when the data is more imbalanced, though it performs better than the baseline regardless of the positive/negative instance ratio.

\subsubsection{Effect of Loss Function w/o Multi-tasking}
Recall that in the multi-task architecture that we have proposed, we use the pairwise ranking loss for the multiple classification task of relation classification and use cross-entropy loss for the binary classification task of relation identification. 

We can, however, use cross-entropy in relation classification as well. To see how the choice of loss function affects performance in different scenarios, we switched ranking loss to cross-entropy loss or simply added cross-entropy loss in the relation classification task, and evaluated the Baseline+Tag model w/o multi-task learning, using the Chinese corpus of ACE 2005. The results are summarized in Table \ref{experiment_loss}, from which we can see:

\begin{enumerate}
	\item When doing a single task of relation classification, the model has higher precision and lower recall with cross-entropy loss, but has lower precision and higher recall with ranking loss; the F1-scores do not differ much. This suggests that for doing relation classification only, the choice of loss function seems not to matter too much.
	
	\item Multi-task learning helps, regardless of the loss function used in relation classification.
	
	\item When we use cross-entropy loss and ranking loss at the same time for relation classification, without multi-tasking, the F1-score only increases slightly from 61.4\% to 61.7\%. But when cross-entropy is applied to another related task---relation identification, with multi-tasking, the F1-score increases from 61.4\% to 62.9\% with an absolute increase of 1.5\%. This suggests that the effectiveness of our multi-task architecture mostly comes from the introduction of relation identification, and this binary classification task does help with the data imbalance problem, corroborating our motivation stated in Section \ref{sec:introduction}.
	
	\item In the same multi-tasking scenario, using ranking loss in relation classification is better than using cross-entropy loss (62.9\% v.s. 62.0\%), with an absolute increase of 0.9\% in F1-score. Note that cross-entropy loss is already used in relation identification. This suggests that the diversity that comes with ranking loss can improve performance.
	
\end{enumerate}

%% file: Literature.tex
\section{Related work}\label{sec:Related work}

\citet{liu2013convolution} were the first to adopt deep learning for relation extraction. They proposed to use a CNN to learn features automatically without using handcraft features. 
\citet{zeng2014relation} also employed CNN to encode the sentence, using additional lexical features to word embeddings.
Their biggest contribution is the introduction of position embeddings.
\citet{zeng2015distant} proposed a model named Piecewise Convolutional Neural Networks (PCNN) in which each convolutional filter $p_i$ is divided into three segments $(p_{i1}, p_{i2}, p_{i3})$ by head and tail entities, and the max-pooling operation is applied to these three segments separately. \citet{santos2015classifying} also used CNN but proposed a new pairwise ranking loss function to reduce the impact of negative instances. \citet{lin2016neural} used CNN with a sentence-level attention mechanism over multiple instances to reduce noise in labels. 


RNN is also widely used in relation extraction. \citet{miwa2016end} used LSTM and tree structures for relation extraction task. Their model is composed of three parts: an embedding layer to encode the input sentence, a sequence layer to identify whether a word is an entity or not, and a dependency layer for relation extraction. \citet{zhou2016attention} used BiLSTM and attention mechanism to improve the model's performance. \citet{she2018distant} proposed a novel Hierarchical attention-based Bidirectional Gated recurrent neural network (HBGD) integrated with entity descriptions to mitigate the problem of having wrong labels and enable the model to capture the most important semantic information. 

Entity background knowledge also contains important information for relation extraction. To capture such information, \citet{ji2017distant} and \citet{she2018distant} extracted entity descriptions from Freebase and Wikipedia and used an encoder to extract features from these descriptions. \citet{he2018see} used a dependency tree to represent the context of entities and transformed the tree into entity context embedding using tree-based GRU.

Unlike most existing works which only consider a single entity pair in a sentence, \citet{christopoulou2019walk} considered multiple entity pairs in a sentence simultaneously and proposed a novel walk-based model to capture the interaction pattern among the entity pairs. \citet{su2017global} pointed out that the global statistics of relations between entity pairs are also useful, and proposed to construct a relation graph and learn relation embeddings to improve the performance of relation extraction.

Several studies are motivated to mitigate the effect of wrong labels \cite{lin2016neural,she2018distant,Qin2018DSGAN}, and \citet{Qi2014Joint} proposed to jointly extract entity mentions and relations. This is not the focus of our paper.

%% file: Conclusion.tex
\section{Conclusion}\label{sec:conclusion}

In this paper, we focus on the relation extraction task with an imbalanced corpus. To mitigate the problem of having too many negative instances, we propose a multi-task architecture which jointly trains a model to perform the relation identification task with cross-entropy loss and the relation classification task with ranking loss. Moreover, we introduce the embeddings of character-wise/word-wise BIO tag from the named entity recognition task to enrich the input representation. Experiment results on ACE 2005 Chinese and English corpus show that our proposed approach can successfully address the data imbalance problem and significantly improve the performance, outperforming the state-of-the-art models in terms of F1-score. Particularly, we find BIO tag embeddings very effective, which we believe could be used as a general part of character/word representation.